\documentclass[runningheads,a4paper]{llncs}

\usepackage{amssymb}
\usepackage{graphicx}
\usepackage{subfig}

\usepackage[T2A,T1]{fontenc}
\usepackage[utf8]{inputenc} 
\usepackage[russian,english]{babel} 

\usepackage{url}
\urldef{\mailsa}\path|{alik.kirillovich, onevzoro, elipachev}@gmail.com|
\newcommand{\keywords}[1]{\par\addvspace\baselineskip
\noindent\keywordname\enspace\ignorespaces#1}

\begin{document}

\mainmatter  

\title{OntoMath${}^{\mathbf{PRO}}$ 2.0 Ontology:\\
Updates of the Formal Model}
\titlerunning{OntoMath${}^{\mathbf{PRO}}$ 2.0 Ontology}
\author{Alexander Kirillovich${}^{1,2}$ \and
  Olga Nevzorova${}^1$ \and Evgeny Lipachev${}^1$}
\authorrunning{Kirillovich, Nevzorova, Lipachev}
\institute{ ${}^1$Kazan Federal University, Kazan, Russia\\
 ${}^2$Joint Supercomputer Center of the Russian Academy of Sciences, Kazan, Russia\\
\mailsa\\
}

\toctitle{OntoMath${}^{\mathbf{PRO}}$ 2.0 Ontology}
\tocauthor{Kirillovich,  Nevzorova, Lipachev}

\maketitle

\begin{abstract} 
This paper is devoted to the problems of ontology-based mathematical knowledge management and representation.
The main attention is paid to the development of a formal model for the representation of mathematical statements
in the Open Linked Data cloud.
The proposed model is intended for applications that extract mathematical facts from natural language mathematical texts and represent these facts as Linked Open Data.
The model is used in development of a new version of the OntoMath${}^{\mathrm{PRO}}$ ontology of professional mathematics is described.
OntoMath${}^{\mathrm{PRO}}$ underlies a semantic publishing platform, that takes as an input a collection of mathematical papers
in \LaTeX{} format and builds their ontology-based Linked Open Data representation.
The semantic publishing platform, in turn, is a central component of OntoMath digital ecosystem, an ecosystem of ontologies,
text analytics tools,
and applications for mathematical knowledge management, including semantic search for mathematical formulas and
a recommender system for mathematical papers.
According to the new model, the ontology is organized into three layers: a foundational ontology layer,
a domain ontology layer and a linguistic layer.
The domain ontology layer contains language-independent math concepts.
The linguistic layer provides linguistic grounding for these concepts,
and the foundation ontology layer provides them with meta-ontological annotations.
The concepts are organized in two main hierarchies: the hierarchy of objects and the hierarchy of reified relationships.
\footnote{Please, cite as:
Alexander Kirillovich, Olga Nevzorova, and Evgeny Lipachev. OntoMath${}^{\mathbf{PRO}}$ 2.0 Ontology: Updates of Formal Model // Lobachevskii Journal of Mathematics, 2022, Vol.~43, No.~12, pp.~3504--3514.
\url{https://doi.org/10.1134/S1995080222150136}
}
\end{abstract}



\section{Introduction}
This paper is devoted to a problem in the field of ontology-based mathematical knowledge management and representation,
i.e. representation of mathematical statements in the Linked Open Data (LOD) cloud.

There are several formalisms for mathematical knowledge representation~\cite{Kali}.
OpenMath \cite{Buswell}, \cite{OpenMath} and Content MathML \cite{MathML} are used to represent mathematical statements,
while OMDoc \cite{OMDoc} is used to represent complex semiformal mathematical documents.
However, these formalisms are not LOD-native and require adaptation to be used in LOD.
In fact, there are projects for encoding OpenMath documents as LOD-integrated RDF datasets
\cite {OpenMath-RDF}, \cite{OpenMath-RDF+}.
These encodings however represent math statements only indirectly: they assert statements about OpenMath documents,
not the math statements themselves.

We propose a formal model for representing mathematical statements
in the Linked Open Data cloud in direct way. This model has been
tested in development of OntoMath${}^{\mathrm{Edu}}$, an
experimental educational mathematical ontology
~\cite{OntoMathEdu-Towards}--\cite{OntoMathEdu++}. Now model is used in developing of the new
version of OntoMath${}^{\mathrm{PRO}}$, an ontology of professional
mathematics.

The new version of OntoMath${}^{\mathrm{PRO}}$ ontology is intended
to be used for extracting mathematical statements from natural
language mathematical texts and representing extracted statements as
Linked Open Data (LOD). LOD have value in themselves~\cite{Lange},
and can be used for navigation, querying, aggregation,
etc~\cite{Dev} -- \cite{BigMath+}.

Additionally, via the OpenDreamKit project and Math-in-the-Middle
ontology~\cite{Deh}, a LOD representation can then be converted to
formats of computer algebra systems. So, for example,
OntoMath${}^{\mathrm{PRO}}$ can be used to parse a mathematical task
in a natural language, and automatically solve this task by a
computer algebra system.

Although the first version of OntoMath${}^{\mathrm{PRO}}$
ontology~\cite{LJM-2014}, \cite{onto-2014} has proven to be
effective in several specialized services, its architecture has a
number of limitations that impede its using for the intended
purpose. In this regard, we started a project for developing the new
major version of the ontology based on the new architecture,
designed to tackle these problems.

The rest of the paper is organized as following. In Section 2 we
describe the first version of the ontology and outline its
restrictions. In Section~2 we describe the architecture of the new
version of ontology under development. In Section 3 we discuss the
task of population the ontology by new concepts. In Conclusions, we
summarize the current status of the project and the directions of
future work.

\section{Ontological representation of math knowledge in OntoMath${}^{\mathrm{PRO}}$ 1.0 ontology}
In this section we briefly describe the first version of  the
OntoMath${}^{\mathrm{PRO}}$ 1.0 ontology and outline its
restrictions.

OntoMath${}^{\mathrm{PRO}}$ 1.0 ontology is organized in two hierarchies:
the hierarchy of fields of mathematics and the hierarchy of objects of mathematical knowledge.
The ontology defines five types of relationships between concepts.
The concept description contains a name in Russian and English, a definition, links to other concepts
and external resources from the Linked Open Data cloud.

The ontology can be used to represent individual mathematical
objects as instances of the classes from the hierarchy of objects.

The ontology is expressed by OWL DL (Web Ontology Language)
formalism which is based on a description logic~\cite{owl-dl}.

OntoMath${}^{\mathrm{PRO}}$ underlies a semantic publishing
platform~\cite{Nevz}, that takes as an input a collection of
mathematical papers in \LaTeX{} format and builds their
ontology-based Linked Open Data representation. The semantic
publishing platform, in turn, is a central component of OntoMath
digital ecosystem~\cite{OntoMath}, \cite{toronto}, an ecosystem of
ontologies, text analytic tools, and applications for mathematical
knowledge management, including semantic search for mathematical
formulas~\cite{RPC} and a recommender system for mathematical
papers~\cite{DAN-2016}.

Although the OntoMath${}^{\mathrm{PRO}}$ 1.0 has proven to  be
effective in several specialized services, its architecture has a
number of limitations. Examples of the most significant restrictions
are
\begin{enumerate}
    \item The existing version of the ontology contains a large number of general concepts, but a rather poor set of relationships between them.
    \item The existing version of the ontology does not distinguish type concepts and role concepts; accordingly,
there are no relations between roles and types.
    \item Linguistic information about concepts is expressed using simple rdf labels that do not contain information
about their internal structure and linguistic properties.
    \item The existing version of the ontology does not contain individuals, but is based solely on the representation of classes.
\end{enumerate}

Such an ontology is suitable for extracting and representing individual mathematical objects,
but not relationships between mathematical objects and mathematical facts.

\section{Ontology-based representation of mathematical statements in Linked Open Data}

The key problem that the new version of the ontology  is intended to
address is representation of mathematical statements as Linked Open
Data. Thus the ontology determines a functions that translate a
mathematical sentence expressed in First-order logic (FOL) to a
LOD-compatible RDF graph. For the obtained RDF translation preserves
the meaning of the source FOL sentence, the source sentence and the
obtained RDF graph must be semantically equivalent. From the
model-theoretic point of view, it means that the source statement
and its translation share the same set of models (in this section we
use the term `model' in the technical sense of the model theory,
where it means an interpretation satisfying a sentence or RDF
graph).

The full coincidence of the models is not possible  however. First,
FOL and RDF have different semantics \cite{RDF}: FOL interpretations
contain arbitrary $n$-ary predicates while a RDF/OWL interpretations
contains only unary and binary, FOL interpretations support
functional terms while RDF/OWL interpretations do not, and so on.
Thus, we can only speak on a some kind of isomorphism between models
of a FOL sentence and its RDF translation. Second, FOL is of more
expressive power then RDF/OWL and so the models of a FOL statement
may be only a subset of the models of its RDF translation.

State the condition that the translation function must satisfy in more formal way.

Let $o$ is the graph of the ontology.

Let $S_{FOL}$ is the set of all FOL sentences.

Let $S_{RDF}$ is the set of all RDF graphs.

Let $*: S_{FOL} \rightarrow S_{RDF}$ is a partial function, that
translate  FOL sentences to RDF graphs.

Let $INT_{FOL}$ is a set of all FOL interpretations.

Let $INT_{RDF}$ is a set of all RDF/OWL interpretations.

Let $t: INT_{FOL} \rightarrow INT_{RDF}$ is a mapping from FOL
interpretations to RDF/OWL interpretations. Define $t$ on a set
$INTS_{FOL}  \subset INT_{FOL}$ of FOL interpretation as: $t
(INTS_{FOL}) = \{t (i) \mid i \in INTS_{FOL}\}$.

Let $M_{FOL} (s)$ is a set of the models of a FOL sentence $s$. Let
$M_{RDF} (g)$ is a set of the models of a RDF graph $g$.

Given a FOL sentence $s$ expressed in terms of a FOL theory $T$, the
translation function $*$ must satisfy the following condition
$$t (M_{FOL} (T \cup \{s\})) \subset M_{RDF} (*(s) \cup o).$$

\begin{figure}[t]
\includegraphics[width=1.00\textwidth]{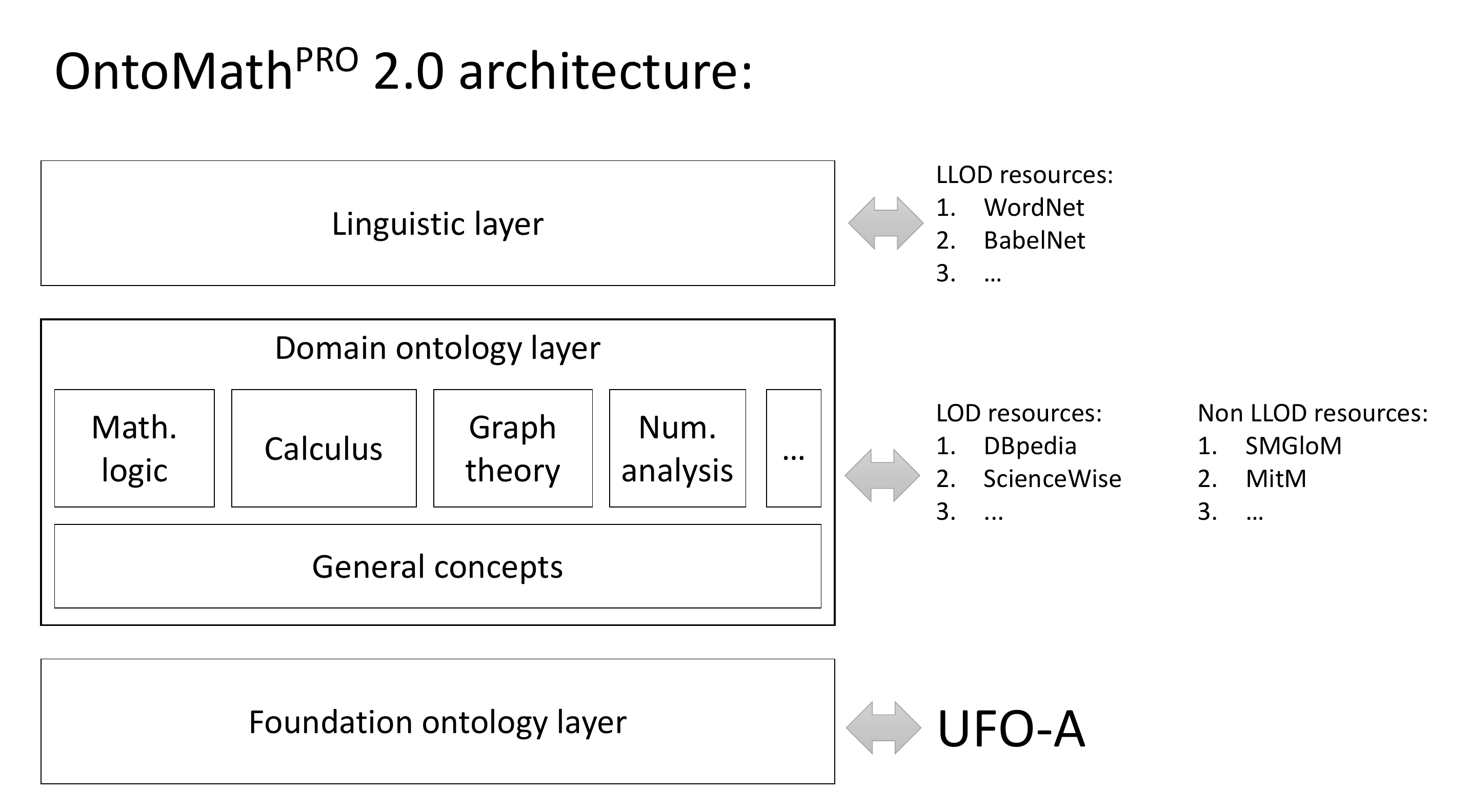}
\caption{OntoMath${}^{\mathrm{PRO}}$ ontology structure (see \cite{dan-2022}) }\label{fig-arcitecture}
\end{figure}

\section{Formal model of OntoMath${}^{\mathrm{PRO}}$ 2.0 ontology}
In this section we describe the model for the new version of  OntoMath${}^{\mathrm{PRO}}$~2.0 ontology.

\begin{figure}[t]
\centering
\includegraphics[width=0.85\textwidth]{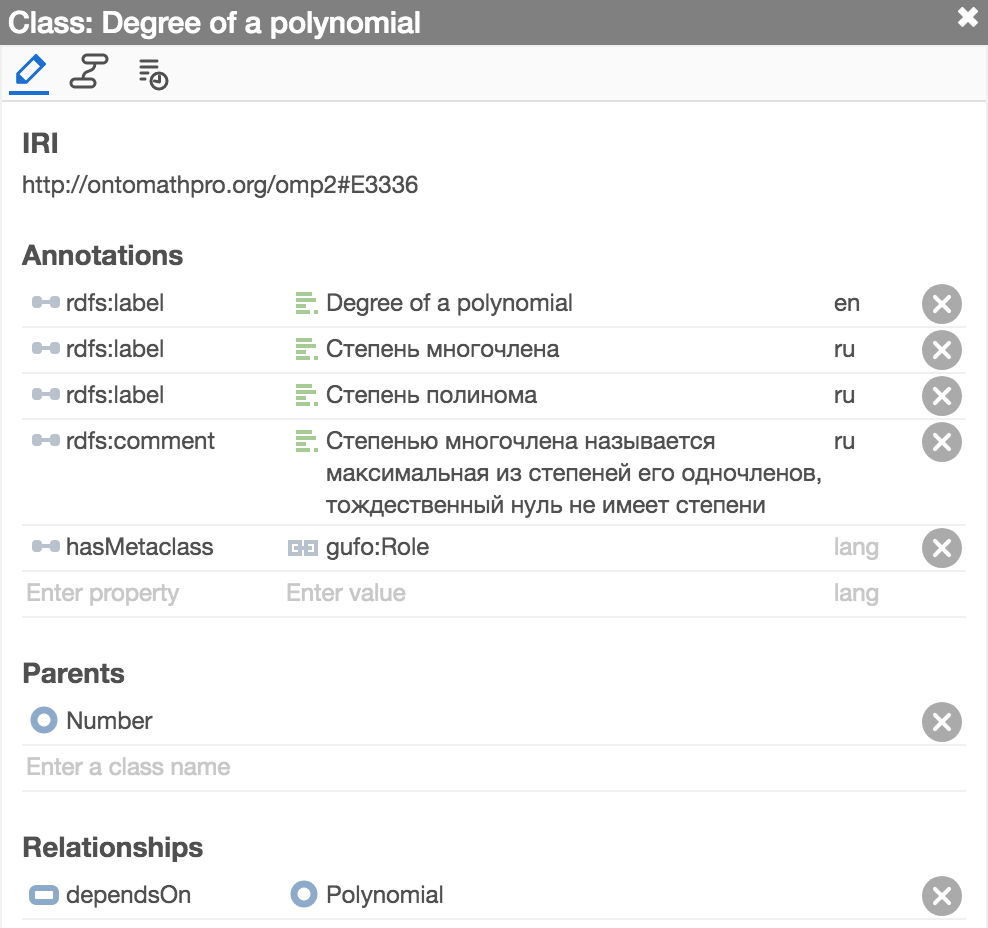}
\caption{Description of the role concept \textit{Degree of a polynomial} in WebProt\'{e}g\'{e} }\label{fig-concept}
\end{figure}

As the first version,  OntoMath${}^{\mathrm{PRO}}$ 2.0 is expressed
by  the OWL~2 DL formalism. The examples below are expressed by
Turtle serialization of RDF and OWL.

According to the model, OntoMath${}^{\mathrm{PRO}}$ 2.0 is organized in three layers:
\begin{enumerate}
    \item \textbf{Domain ontology layer}, which contains language-independent math concepts.
    \item \textbf{Linguistic layer}, containing multilingual lexicons,
    that provide linguistic grounding of the concepts from the domain ontology layer.
    \item \textbf{Foundational ontology layer}, that provides the concepts with
    meta-ontological annotations.
\end{enumerate}

This three-layered structure is represented at Figure~\ref{fig-arcitecture}.

The domain ontology layer is organized in two main hierarchies:
the hierarchy of objects and the hierarchy of reified relationships.

Figure~\ref{fig-concept} depicts a description of the role concept \textit{Degree of
a polynomial} in the WebProt\'{e}g\'{e} editor. The description of
contains the names of the concept in Russian and English, the
metaontological annotation, the parent concept and a relation with
the type concept \textit{Polynomial}.

\begin{figure}[t]
\centering
\includegraphics[width=1.00\textwidth]{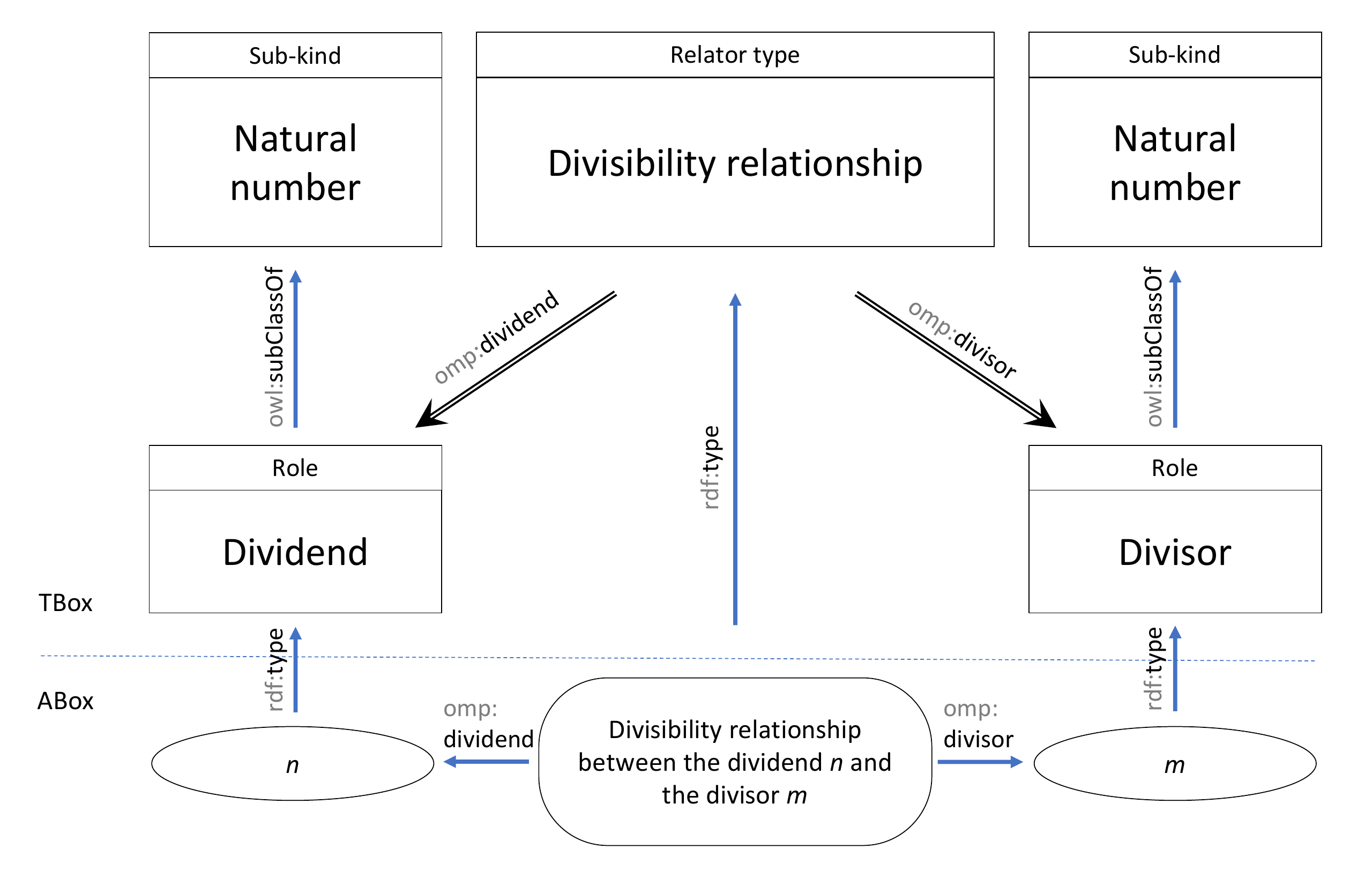}
\caption{An example of a materialized relationship,
and its instance corresponding to the ``The number $m$ divides the number
$n$'' statement (see \cite{dan-2022}).}
\label{fig-relationship}
\end{figure}

There are two meta-ontological types of the concepts: kinds and
roles. A kind is a concept that is rigid and ontologically
independent~\cite{UFO}, \cite{Guizz}. So, for example, the
\textit{Integer} concept is a kind, because any integer is always a
triangle, regardless of its relationship with other objects. A role
is a concept that is anti-rigid and ontologically
dependent~\cite{UFO}, \cite{Guizz}. An object can be an instance of
a role class only by virtue of its relationship with another object.
So, for example, the \textit{Degree of a polynomial} concept  is a
role, since any integer is a degree of a polynomial not by itself,
but only in relation to a certain polynomial. Any role concept is a
subclass of some kind concept. For example, the \textit{Degree of a
polynomial} role concept is a subclass of \textit{Natural number}
kind concept.

Relations between concepts are represented in ontology in a reified
form, i.e., as concepts, not as object properties (such
representation fits the standard onto-logical pattern for
representing $N$-ary relation with no distinguished
participant~\cite{Noy}, but is applied to binary relations too).
Thus, the relationships between concepts are first-order entities,
and can be a subject of a statement.

Reified relationships are linked to their participants by
\textit{has argument} object properties and their subproperties.

Figure~\ref{fig-relationship} shows one of the relations, represented by the
\textit{Divisibility relationship} concept. This relation is linked
to its participants, represented by \textit{Dividend} and
\textit{Divisor} role concepts. These roles, in turn, are defined as
subclasses of the \textit{Natural number} kind concept. The bottom
of the figure depicts an instance of this relation, namely the
\textit{Divisibility relationship between the dividend $n$ and the
divisor $m$}, that binds the natural number $n$ and the natural
number $m$ (see also~\cite{dan-2022}). This instance is a
representation of natural language statement ``The number $m$
divides the number $n$''.

The mappings between ontology concepts and corresponding natural language statements are defined at the linguistic
level of the ontology.

Construction with a reified relationship is a RDF translation of the
mathematical sentence that is expressed in FOL  as an atomic
formula. Translation of a formula from FOL to RDF is defined as
follows.

Let $T$ is a mathematical theory, $RNames_{math}$ is a  set of
predicate letters of  $T$ and $Const_{math}$ is a set of constants
of $T$.

Let $RNames_{owl}$ is a set of URIs of classes  from the hierarchy
of reified relationships of the OntoMath${}^{\mathrm{PRO}}$ ontology
and $Const_{owl}$ is a set of URIs denoting mathematical objects.

Let $pmap: RNames_{math} \rightarrow RNames_{owl}$ and   $cmap:
Const_{math} \rightarrow Const_{owl}$.

A FOL sentence $R(c_1, ..., c_n)$, where  $R \in RNames_{math}$,
$c_1, ..., c_n \in  Const_{math}$ is translated to the following RDF
graph
\begin{verbatim}
@prefix rdf: <http://www.w3.org/1999/02/22-rdf-syntax-ns#> .
@prefix rdfs: <http://www.w3.org/2000/01/rdf-schema#> .
@prefix owl: <http://www.w3.org/2002/07/owl#> .
@prefix omp: <http://ontomathpro.org/omp2#> .

_:rel rdf:type rmap(R).
_:rel omp:hasArgument cmap (c1).
...
_:rel omp:hasArgument cmap (cn).
\end{verbatim}

\begin{figure}[t]
\centering
\includegraphics[width=0.70\textwidth]{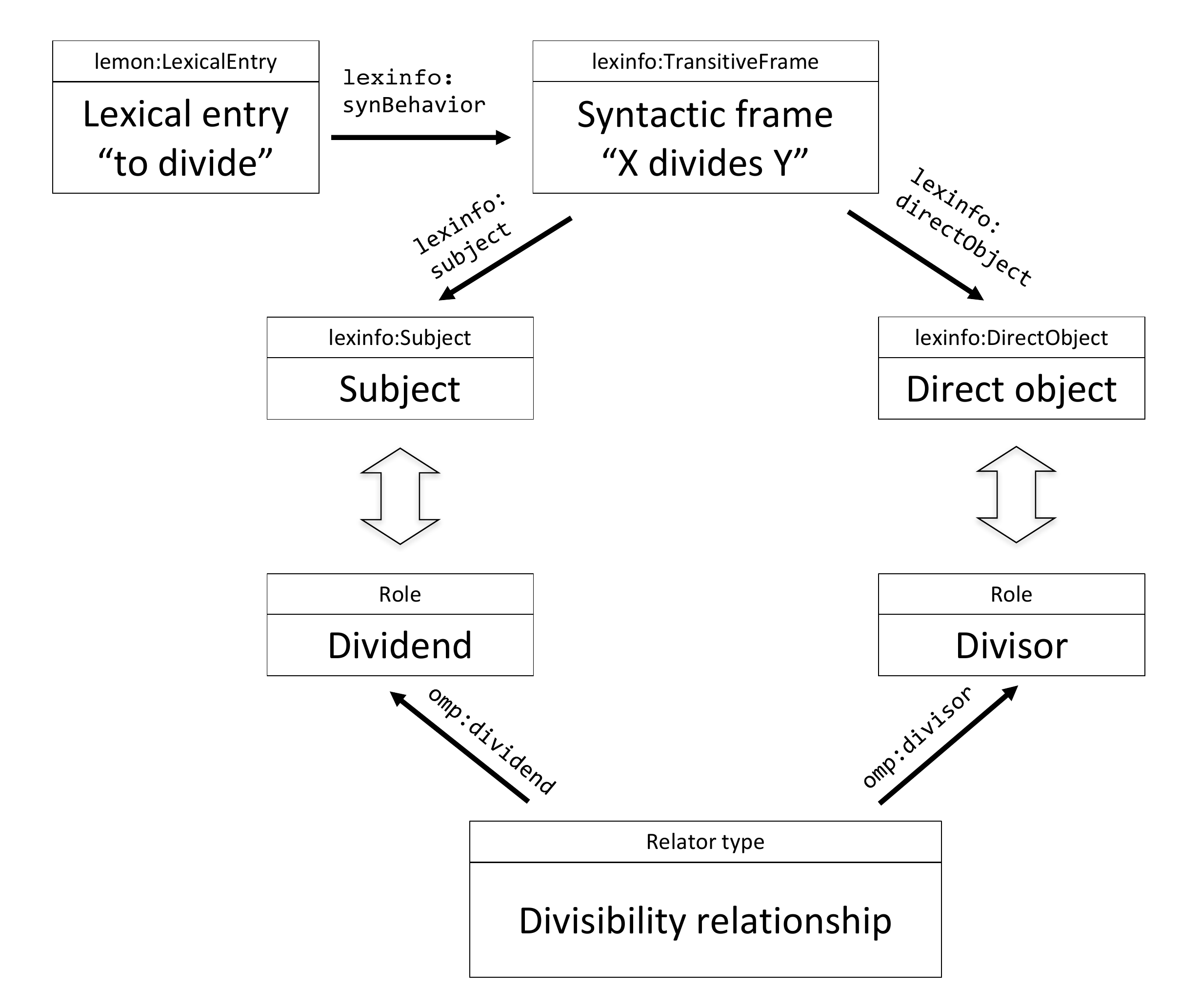}
\caption{Syntactic frame for the ``divide'' verb and its mapping to \emph{Divisibility} relationship. The subject and  the object syntactic arguments of the frame
are mapped to \emph{Dividend} and \emph{Divisor} arguments of the
the relationship.} \label{fig-frame}
\end{figure}

The mappings between ontology concepts  and corresponding natural
language statements are defined at the linguistic level of the
ontology.

The linguistic layer contains multilingual  lexicons, that provide
linguistic grounding of the concepts from the domain ontology layer.
Currently we are developing the lexicons for Russian and English. A
lexicon consists in
\begin{itemize}
    \item Lexical entries, denoting mathematical concepts.
    Examples of lexical entries are
    ``number'', ``prime number'', ``degree of a polynomial'',
    ``to intersect", etc.
    \item Forms of lexical entries (in different numbers, cases, tenses, etc).
    \item Syntactic trees of multi-word lexical entries.
    \item Syntactic frames of lexical entries. A syntactic frame represents the syntactic behavior of a predicate,
defining the set of syntactic arguments this predicate requires and
their mappings to ontological entities. For example, a syntactic
frame of the ``to divide'' verb determines that in ``$X$ divides
$Y$'' phrase, the subject $X$ represents the divisor and the  direct object $Y$ represents the dividend (Figures~\ref{fig-frame}, \ref{fig-frame-llod}).
\end{itemize}

\begin{figure}[h!]
\begin{verbatim}
@prefix : <http://ontomathpro.org/lexicons/>.
@prefix omp: <http://ontomathpro.org/omp2#> .
@prefix ontolex: <http://www.w3.org/ns/lemon/ontolex#>.
@prefix synsem: <http://www.w3.org/ns/lemon/synsem#>.
@prefix lexinfo: <http://www.lexinfo.net/ontology/2.0/lexinfo#>.

#The "to divide" verb
:EN-v-divide
  a ontolex:LexicalEntry;
  lexinfo:partOfSpeech lexinfo:verb;
  ontolex:canonicalForm :EN-v-divide-form0;
  synsem:synBehavior :EN-v-divide-frame1;
  ontolex:sense :EN-v-divide-sense1.

#The canonical form of this verb
:EN-v-divide-form0
  a ontolex:Form;
  ontolex:writtenRep "divide"@en.

#A syntactic frame "X dividees Y"
:EN-v-divide-frame1
  a lexinfo:TransitivePPFrame;
  lexinfo:subject :EN-v-divide-frame1-subj;
  lexinfo:directObject :EN-v-divide-frame1-obj.

#The subject of the verb
:EN-v-divide-frame1-subj
  a lexinfo:Subject.
  
#The direct object of the verb
:EN-v-divide-frame1-obj
  a lexinfo:DirectObject.
\end{verbatim}
\caption{Syntactic frame for the ``divide'' verb and its mapping to \emph{Divisibility} relationship in the LLOD format}
\label{fig-frame-llod}
\end{figure}

\begin{figure}[h!]
\ContinuedFloat
\begin{verbatim}

#A lexical sense of the verb, expressing the mapping to ontology
:EN-v-divide-sense1
  a ontolex:LexicalSense, synsem:OntoMap;
  synsem:ontoMapping :EN-v-divide-sense1;
  ontolex:reference omp:Divisibility_relationship;
    
  synsem:submap
    [
    a synsem:OntoMap;
    ontolex:reference  omp:Divisibility_relationship;
    synsem:isA _:Relationship1
    ], [
    a synsem:OntoMap;
    ontolex:reference omp:dividend;
    synsem:subjOfProp _:Relationship1;
    synsem:objOfProp :EN-v-divide-frame1-subj
    ], [
    a synsem:OntoMap;
    ontolex:reference omp:divisor;
    synsem:subjOfProp _:Relationship1;
    synsem:objOfProp :EN-v-divide-frame1-obj
    ], [
    a synsem:OntoMap;
    ontolex:reference omp:Dividend;
    synsem:isA :EN-v-divide-frame1-subj
    ], [
    a synsem:OntoMap;
    ontolex:reference omp:Divisor;
    synsem:isA :EN-v-divide-frame1-pp_at
    ].
\end{verbatim}
\caption{Syntactic frame for the ``divide'' verb and its mapping to \emph{Divisibility} relationship in the LLOD format (continuation)}
\end{figure}

The lexicons are expressed as Linguistic  Linked Open Data (LLOD)
datasets in terms of Lemon ontology~\cite{Lemon-1}, \cite{Lemon-2},
LexInfo~\cite{LexInfo}, OLiA~\cite{OLiA} and PreMOn~\cite{PreMOn}
ontologies (see also~\cite{Lemon+}).

Lemon ontology is used to represent complex lexical resources. This
ontology is based on the international standard ISO 24613:2008
``Language resource management -- Lexical markup framework
(LMF)''~\cite{LMF}. The basic elements of this ontology are
lexicons, lexical entries, forms of a lexical entry, senses of a
lexical entry, and concepts from ontologies of subject domains.

To describe language categories (gender, number, case, tense, direct object, in-direct
object, synonym, antonym, etc.) in Lemon ontology, the external ontology LexInfo is used.
To describe language categories in LLOD, different ontologies are used that are
linked to the ISOcat data category registry, which is the implementation of the
international standard ISO 12620: 2009 ``Terminology and other languages and content resources --
Specification of data categories and management of the Data Category''~\cite{ISO-12620}.

\section{Replenishing the ontology by new concepts}
When, developing the ontology, various sources of mathematical
knowledge are used, including articles from mathematical
journals~\cite{Bir}, \cite{dan-2014}.

Despite the rather significant amount of mathematical terms,  the
replenishment of ontologies is an urgent task. Mathematical
encyclopedias and reference books are the good sources of
replenishment of ontologies. We compared the volume of terminology
from the ontology with the terminological index of the
``Mathematical Handbook for Scientists and Engineers'' by G.~Korn
and T.~Korn~\cite{Korn} based on a specially developed program and
got quite interesting results.

The comparison was carried out only in the fields of mathematics, which are represented in the ontology.
In total, there are 2227 terms in Korn's reference book, of which 791 terms have an intersection with terms
from the ontology. When comparing terms, a cosine measure was used with text preprocessing.

Pre-processing consisted in removing punctuation marks, stop words, and lemmatizing all words used
pymorphy2 library~\cite{polymorphy2}, \cite{Korobov}.
The threshold value for an acceptable measure of similarity was chosen to be $0.7$.

The following situations were identified when comparing terminology
\begin{itemize}
    \item use of incomplete labels in the ontology
    (\textit{Riemann--Stieltjes integral} / \textit{Riemann--Stieltjes probability integral}~\cite{Korn};
    \textit{Cesaro summable series} / s\textit{ummable series by Cesaro method}~\cite{Korn}, etc.);
    \item the comparison scores are high (above the threshold value of $0.7$),
    but actually specific value and general value of term were compared
    (e.g., \textit{Stormer interpolation formula}/\textit{interpolation formula}~\cite{Korn};
    \textit{Gaussian interpolation formula}/\textit{interpolation formula}~\cite{Korn};
    \textit{Adams interpolation formula}/\textit{Adams formula}~\cite{Korn}).
\end{itemize}

In the last example, professional review is required to match the terms.
Thus, replenishment of the ontology is a very non-trivial task.

\section{Conclusions}
In this article, we discussed the previous version of the
OntoMath${}^{\mathrm{PRO}}$   ontology and the reasons why it became
necessary to improve the formal model for representing mathematical
knowledge.

We have described the new formal model that underlies the new
version  of OntoMath${}^{\mathrm{PRO}}$ ontology of professional
mathematics. According to this formal model, the ontology is
organized into three layers: a foundational ontology layer, a domain
ontology layer and a linguistic layer. The developed formal model
allows representing mathematical statements in Open Linked Data
cloud.

Our work on the new version of the ontology is determined  by the
proposed model and undergoing by the following steps
\begin{enumerate}
    \item Internal verification and correction the taxonomy of the ontology.
    \item Providing the concepts with meta-ontological annotations.
    \item Development of materialized relationships.
    \item Development of linguistic layer.
    \item Replenishing the ontology by new concepts.
    \item External verification of the new version of ontology in such tasks as
    information extraction and semantic search.
\end{enumerate}

On the current stage, we have performed the internal  verification
of the taxonomy and provided the concepts with meta-ontological
annotations. Currently, we are working on development of reified
relationships and have developed several experimental ones.

{\bf Funding.} The work was funded by Russian Science Foundation
according to the research project no. 21--11--00105.

%
%


\begin{thebibliography}{99}

\bibitem{Kali}
C.~Kaliszyk and F.~Rabe, ``A Survey of Languages for Formalizing
Mathematics,'' In: C.~Benzm\"{u}ller and B.~Miller B. (Eds.) CICM
2020. Lecture Notes in Artificial Intelligence \textbf{12236},
138--156 (2020). \url{https://doi.org/10.1007/978-3-030-53518-6_9}.

\bibitem{Buswell}
S.~Buswell, O.~Caprotti, D.P.~Carlisle, M.C.~Dewar, M.~Gaetano, M.~Kohlhase,
J.H.~Davenport, P.D.F.~Ion, and T.~Wiesing(Eds.).
\textit{The OpenMath Standard. Version:
2.0r2} (The OpenMath Society, July 2019).
\url{https://openmath.org/standard/om20-2019-07-01/omstd20.html}.

\bibitem{OpenMath}
M.~Kohlhase, and F.~Rabe, ``Semantics of OpenMath and MathML3,''
Mathematics in Computer Science \textbf{6} (3), 235--260 (2012).
\url{https://doi.org/doi:10.1007/s11786-012-0113-x}.

\bibitem{MathML}
D.~Carlisle, P.~Ion, and  R.~Miner (Eds.),
\textit{Mathematical Markup Language
(MathML). Version 3.0. 2nd Edition} (W3C Recommendation, 10 April
2014). \url{https://www.w3.org/TR/MathML3/}.

\bibitem{OMDoc}
M.~Kohlhase,
\textit{OMDoc -- An Open Markup Format for Mathematical Documents [version 1.2].} Lecture Notes in Artificial Intelligence \textbf{4180} (Springer, 2006).

\bibitem{OpenMath-RDF}
K.~Wenzel,
``OpenMath-RDF: RDF encodings for OpenMath objects and Content Dictionaries,''
 31th OpenMath Workshop at CICM 2021. \url{https://easychair.org/publications/preprint_open/XQwn}

\bibitem{OpenMath-RDF+}
K.~Wenzel and H.~Reinhardt, ``Mathematical Computations for Linked
Data Applications with OpenMath,''
CEUR Workshop Proceedings \textbf{921}, 38--48 (2012).

\bibitem{OntoMathEdu-Towards}
A.~Kirillovich, O.~Nevzorova, M.~Falileeva, E.~Lipachev, and L.~
Shakirova, ``OntoMathEdu: Towards an Educational Mathematical
Ontology,'' In: E.~Brady et al. (Eds.), Workshop Papers at 12th
Conference on Intelligent Computer Mathematics (CICM-WS 2019),
Prague, Czech Republic, 8--12 July 2019, CEUR Workshop Proceedings
\textbf{2634}, 1--10 (2020).
\url{http://ceur-ws.org/Vol-2634/WiP1.pdf}.

\bibitem{OntoMathEdu+}
A.~Kirillovich, O.~Nevzorova, M.~Falileeva, E.~Lipachev, and L.~
Shakirova, ``OntoMathEdu: A Linguistically Grounded Educational
Mathematical Ontology,'' Lecture Notes in Computer Science
\textbf{12236}, 157--172 (Springer, 2020).
\url{https://doi.org/10.1007/978-3-030-53518-6_10}.

\bibitem{OntoMathEdu++}
A.~Kirillovich, M.~Falileeva, O.~Nevzorova, E.~Lipachev, A.~Dyupina,
and L.~Shakirova, ``Prerequisite Relationships of the OntoMathEdu
Educational Mathematical Ontology,'' In: J. C.~Figueroa-Garc\'{\i}a,
Y.~D\'{\i}az-Gutierrez, E. E.~Gaona-Garc\'{\i}a, A.
D.~Orjuela-Ca\~{n}\'{o}n (Eds.) Applied Computer Sciences in
Engineering. WEA 2021. Communications in Computer and Information
Science \textbf{1431}, 517--524 (Springer, Cham, 2021).
\url{https://doi.org/10.1007/978-3-030-86702-7_44}

\bibitem{Lange}
C.~Lange, ``Ontologies and languages for representing mathematical
knowledge on the Semantic Web,'' Semantic Web.  \textbf{4} (2),
119--158 (2013). \url{https://doi.org/10.3233/SW-2012-0059}.

\bibitem{Dev}
{\it Developing a 21st Century Global Library for Mathematics}
(Research, DC: The National Academies Press, 2014).
\url{https://doi.org/10.17226/18619}.

\bibitem{BigMath}
J.~Carette, W. M.~Farmer, M.~Kohlhase, and F.~Rabe, ``Big Math and
the One--Brain Barrier A Position Paper and Architecture Proposal,''
arXiv:1904.10405 [cs.MS] (2019).
\url{https://doi.org/10.48550/arXiv.1904.10405}.

\bibitem{BigMath+}
J.~Carette, W. M.~Farmer, M.~Kohlhase, and F.~Rabe, ``Big Math and
the One-Brain Barrier: The Tetrapod Model of Mathematical
Knowledge,'' Math Intelligencer \textbf{43}, 78--87 (2021).
\url{https://doi.org/10.1007/s00283-020-10006-0}.

\bibitem{Deh}
P.-O.~Dehaye et al., ``Interoperability in the Open-DreamKit
Project: The Math-in-the-Middle Approach,'' In: M.~Kohlhase, M.~Johansson, B.~Miller, L.~de Moura, F.~Tompa,
(Eds.) Proc. of the 9th International Conference on Intelligent
Computer Mathematics (CICM 2016). Lecture Notes in Computer Science
\textbf{9791}, 117--131 (Springer, Cham, 2016).
\url{https://doi.org/10.1007/978-3-319-42547-4_9}.


\bibitem{LJM-2014}
A. M.~Elizarov, A. V.~Kirillovich, E. K.~Lipachev, O. A.~Nevzorova,
V. D.~Solovyev, and N. G.~Zhiltsov, ``Mathematical Knowledge
Representation: Semantic Models and Formalisms,''  Lobachevskii J.
Math. \textbf{35} (4), 347--353 (2014).
\url{https://doi.org/10.1134/S1995080214040143}.

\bibitem{onto-2014}
O.~Nevzorova, N.~Zhiltsov, A.~Kirillovich, and E.~Lipachev,
``OntoMathPRO Ontology: A Linked Data Hub for Mathematics,''
Communications in Computer and Information Science \textbf{468},
105--119 (Springer, Cham, 2014).
\url{https://doi.org/10.1007/978-3-319-11716-4_9}.

\bibitem{owl-dl}
OWL Web Ontology Language Guide. W3C Recommendation 10 February 2004.
URL:
\url{https://www.w3.org/TR/owl-guide/}.

\bibitem{Nevz}
O.~Nevzorova at al., ``Bringing Math to LOD: A Semantic Publishing
Platform Prototype for Scientific Collections in Mathematics,''
Lecture Notes in Computer Science  \textbf{8218}, 379--394
(Springer, Berlin, Heidelberg, 2013).
\url{https://doi.org/10.1007/978-3-642-41335-3_24}.

\bibitem{OntoMath}
A.~Elizarov, A.~Kirillovich, E.~Lipachev,  and O.~Nevzorova,
``Digital Ecosystem OntoMath: Mathematical Knowledge Analytics and
Management,'' Communications in Computer and Information Science
\textbf{706},  33--46 (Springer, Cham, 2017).
\url{https://doi.org/10.1007/978-3-319-57135-5_3}.

\bibitem{toronto}
A. M.~Elizarov, N. G.~Zhiltsov,  A. V. Kirillovich, E. K.~Lipachev,
O. A.~Nevzorova,  and V. D.~Solovyev,  ``The OntoMath Ecosystem:
Ontologies and Applications for Math Knowledge Management,'' In:
Semantic Representation of Mathematical Knowledge Workshop 5
February 2016.
\url{http://www.fields.utoronto.ca/video-archive/2016/02/
2053-14698}.

\bibitem{RPC}
A.~Elizarov, A.~Kirillovich,  E.~Lipachev, and O.~Nevzorova,
``Semantic Formula Search in Digital Mathematical Libraries,''
Proceedings of the 2nd Russia and Pacific Conference on Computer
Technology and Applications (RPC 2017), pp.~39--43. IEEE (2017).
\url{https://doi.org/10.1109/RPC.2017.8168063}.

\bibitem{DAN-2016}
A. M.~Elizarov, A. V.~Kirillovich, E. K.~Lipachev, A.
B.~Zhizhchenko, and N. G.~Zhil'tsov,  ``Mathematical Knowledge
Ontologies and Recommender Systems for Collections of Documents in
Physics and Mathematics,'' Doklady Mathematics \textbf{93} (2),
231--233 (2016). \url{https://doi.org/10.1134/S1064562416020174}.

\bibitem{RDF}
RDF 1.1 Semantics. W3C Recommendation 25 February 2014, URL:
\url{https://www.w3.org/TR/rdf11-mt/}.

\bibitem{UFO}
G.~Guizzardi et al.,  ``UFO: Unified Foundational Ontology,'' Appl.
Ontol. \textbf{17} (1), 167--210 (2022).
\url{https://doi.org/10.3233/AO-210256}.

\bibitem{Guizz}
G.~Guizzardi,  \textit{Ontological Foundations for Structural Conceptual
Models} (CTIT, Enschede, 2005).


\bibitem{Noy}
N.~Noy and  A.~Rector,  \textit{Defining N-ary Relations on the
Semantic Web. W3C Working Group Note, 12 April 2006}
(W3C)
\url{https://www.w3.org/TR/swbp-n-aryRelations/}.

\bibitem{dan-2022}
A. M.~Elizarov, A. V.~Kirillovich, E. K.~Lipachev, and O.
A.~Nevzorova, ``OntoMath${}^\mathrm{PRO}$: Ontology of Mathematical
Knowledge,'' Doklady Mathematics (2022).
\url{https://doi.org/10.1134/S1064562422700016}.

\bibitem{Lemon-1}
P.~Cimiano,  J. P.~McCrae, and P.~Buitelaar, \textit{Lexicon model for
ontologies}. Final community group report, 10 May 2016.
\url{https://www.w3.org/2016/05/ontolex/}.

%
\bibitem{Lemon-2}
J. P.~McCrae, J.~Bosque-Gil,  J.~Gracia,  P.~Buitelaar, and
P.~Cimiano, ``The OntoLex-Lemon Model: Development and
Applications,'' In: I.~Kosem et al. (Eds.) Proc. of the 5th biennial
conference on Electronic Lexicography (eLex 2017), 587--597. Lexical
Computing CZ (2017).

\bibitem{LexInfo}
P.~Cimano, P.~Buitelaar, J. P.~McCrae, and M.~Sintek, ``LexInfo: A
declarative model for the lexicon-ontology interface,'' J. of Web
Semantics \textbf{9} (1), 29--51 (2011).
\url{https://doi.org/10.1016/j.websem.2010.11.001}.

\bibitem{OLiA}
C.~Chiarcos,
``OLiA -- Ontologies of Linguistic Annotation,'' Semantic Web \textbf{6} (4), 379--386 (2015).
\url{https://doi.org/10.3233/SW-140167}.

\bibitem{PreMOn}
M.~Rospocher, F.~Corcoglioniti, and A.~Palmero Aprosio, ``PreMOn:
LODifing linguistic predicate models,'' Language Resources and
Evaluation \textbf{53}, 499--524 (2019).
\url{https://doi.org/10.1007/s10579-018-9437-8}.

\bibitem{Lemon+}
  P.~Cimiano, C.~Unger, and J.~McCrae,
  \textit{Ontology--Based Interpretation of Natural Language}.
  (Springer, Cham, 2014).
 \url{https://doi.org/10.1007/978-3-031-02154-1}.

\bibitem{LMF}
Language resource management -- Lexical markup framework (LMF). ISO/CD 24613:2006 (Iso.org, 2006).
\url{https://lirics.loria.fr/doc_pub/N330_LMF_rev13_For_CD_Ballot.pdf}.

\bibitem{ISO-12620}
Terminology and other languages and content resources --
Specification of data categories and management of the Data Category. ISO 12620:2009 (Iso.org, 2009).
\url{http://atoll.inria.fr/RNIL/TC37SC4-docs/CD12620-1.pdf}.


\bibitem{OntoClean}
N.~Guarino  and C. A.~Welty, ``An Overview of OntoClean,'' In:
S.~Staab,  and R.~Studer (Eds.) Handbook on Ontologies. Springer,
Berlin, Heidelberg (2009), 201--220.
\url{https://doi.org/10.1007/978-3-540-92673-3_9}.

\bibitem{Bir}
E. V.~Biryaltsev, A. M.~Elizarov,  N.G.~Zhiltsov, E. K.~Lipachev, O. A.~Nevzorova, and V. D.~Solovyev, ``Methods
for analyzing semantic data of electronic collections in
mathematics,'' Autom. Doc. Math. Linguist. \textbf{48}, 81--85
(2014). \url{https://doi.org/10.3103/S000510551402006X}

\bibitem{dan-2014}
A. M.~Elizarov, E. K.~Lipachev, O. A.~Nevzorova, and V. D.~Solovyev,
``Methods and Means for Semantic Structuring of Electronic
Mathematical Documents,'' Doklady Mathematics \textbf{90} (1),
521--524 (2014). \url{https://doi.org/0.1134/S1064562414050275}.

\bibitem{Korn}
G. A.~Korn and T. M.~Korn, \textit{Mathematical Handbook for
Scientists and Engineers: Definitions, Theorems, and Formulas for
Reference and Review} (Dover Publications; Revised edition, 2000).

\bibitem{polymorphy2}
Morphological Analyzer pymorphy2. \url{https://pymorphy2.readthedocs.io/en/stable/}.

\bibitem{Korobov}
M.~Korobov, ``Morphological Analyzer and Generator for Russian and
Ukrainian Languages,'' Analysis of Images, Social Networks and Texts,
320--332 (2015).



\end{thebibliography}
\end{document}